\documentclass[11pt,a4paper]{article}
\usepackage[hyperref]{emnlp-ijcnlp-2019}
\usepackage{times}
\usepackage{latexsym}
\usepackage{graphicx}
\usepackage{tabularx}
\usepackage{xcolor}
\usepackage{multirow}
\usepackage{arydshln}

\usepackage{amsfonts} % for \mathbb command

\usepackage{url}

\usepackage{amsmath}
\usepackage{stmaryrd}

\aclfinalcopy % Uncomment this line for the final submission

\title{Bidirectional Context-Aware Hierarchical Attention Network for Document Understanding}

\author{Jean-Baptiste Remy$^{1,\dagger}$ \hspace{0.5cm} Antoine J.-P. Tixier$^{1,\dagger}$ \hspace{0.5cm} Michalis Vazirgiannis$^{1,2}$\\
     $^1$\'Ecole Polytechnique, France; $^2$AUEB, Greece\\
    {\tt \small \{jean-baptiste.remy,anti5662,michalis.vazirgiannis\}@polytechnique.edu} \\
 $^\dagger$Equal contribution.
    }

\date{2019}

\begin{document}
\maketitle

\begin{abstract}
The Hierarchical Attention Network (HAN) has made great strides, but it suffers a major limitation: at level 1, each sentence is encoded in complete isolation. In this work, we propose and compare several modifications of HAN in which the sentence encoder is able to make context-aware attentional decisions (CAHAN). Furthermore, we propose a bidirectional document encoder that processes the document forwards and backwards, using the preceding and following sentences as context. Experiments on three large-scale sentiment and topic classification datasets show that the bidirectional version of CAHAN outperforms HAN everywhere, with only a modest increase in computation time. While results are promising, we expect the superiority of CAHAN to be even more evident on tasks requiring a deeper understanding of the input documents, such as abstractive summarization. Code is publicly available\footnote{\href{https://github.com/JbRemy/Cahan}{https://github.com/JbRemy/Cahan}}.
\end{abstract}

\section{Introduction}
Recently, hierarchical architectures have become ubiquitous in NLP.
They have been applied to a wide variety of tasks such as language modeling and generation \cite{lin2015hierarchical,li2015hierarchical}, neural machine translation (NMT) \cite{wang2017exploiting}, summarization \cite{celikyilmaz2018deep}, sentiment and topic classification
\cite{tang2015document,yang2016hierarchical}, and spoken language understanding
\cite{raheja2019dialogue,shang2019energy}, to cite only a few examples. All hierarchical architectures capitalize on the
same intuitive idea that the representation of the input text should be learned
in a bottom-up fashion by using a different encoder at each granularity level (e.g., words, sentences, paragraphs), where the encoder at level $l+1$ takes as input the output of the encoder at level $l$.

\begin{figure}[ht]
    \centering
    \includegraphics[width=0.9\columnwidth]{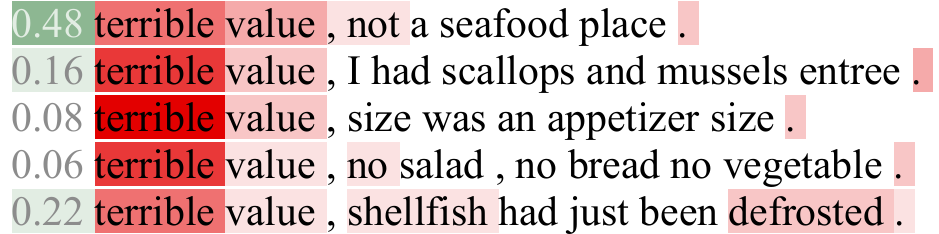}
    \par
    \vspace{0.2cm}
    \includegraphics[width=0.9\columnwidth]{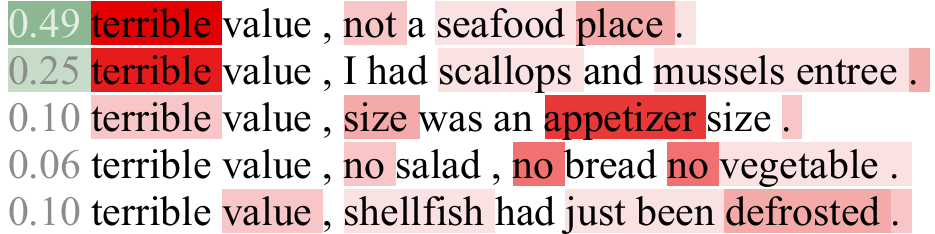}
    \captionsetup{size=small}
    \caption{HAN (top) vs. CAHAN (bottom) on a synthetic example where the same highly negative feature is repeated at the beginning of each sentence to shadow other topics. }\label{fig:example_yelp}
\end{figure}

\begin{figure*}[ht]
    \centering
    \includegraphics[width=0.75\textwidth]{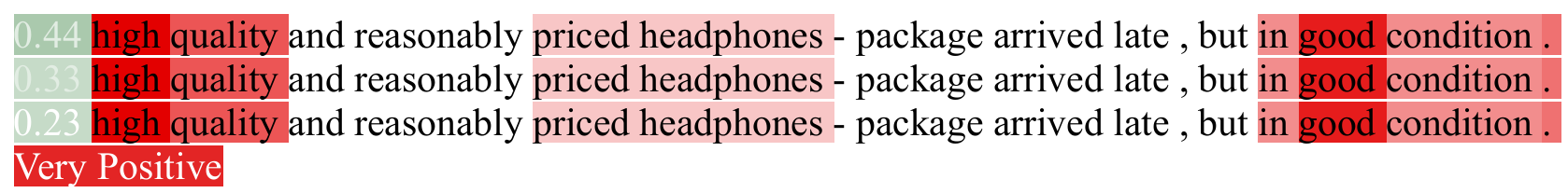}
    \par
    \vspace{0.2cm}
    \includegraphics[width=0.75\textwidth]{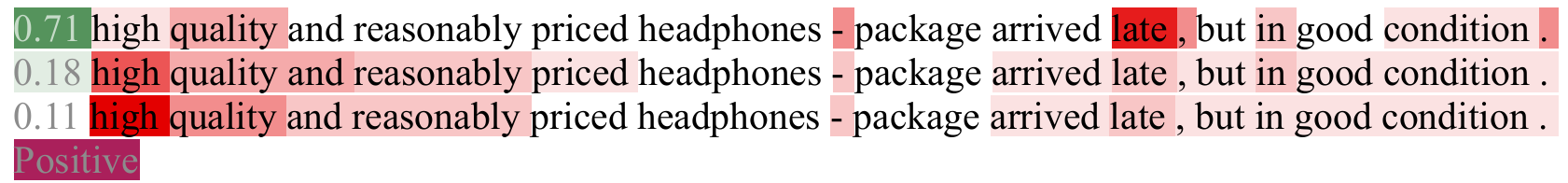}
    \captionsetup{size=small}
    \caption{HAN (top) vs. CAHAN (bottom) on a synthetic example where the same sentence is repeated. }\label{fig:example_amazon}
\end{figure*}

\begin{figure}[ht]
    \centering
    \includegraphics[width=0.9\columnwidth]{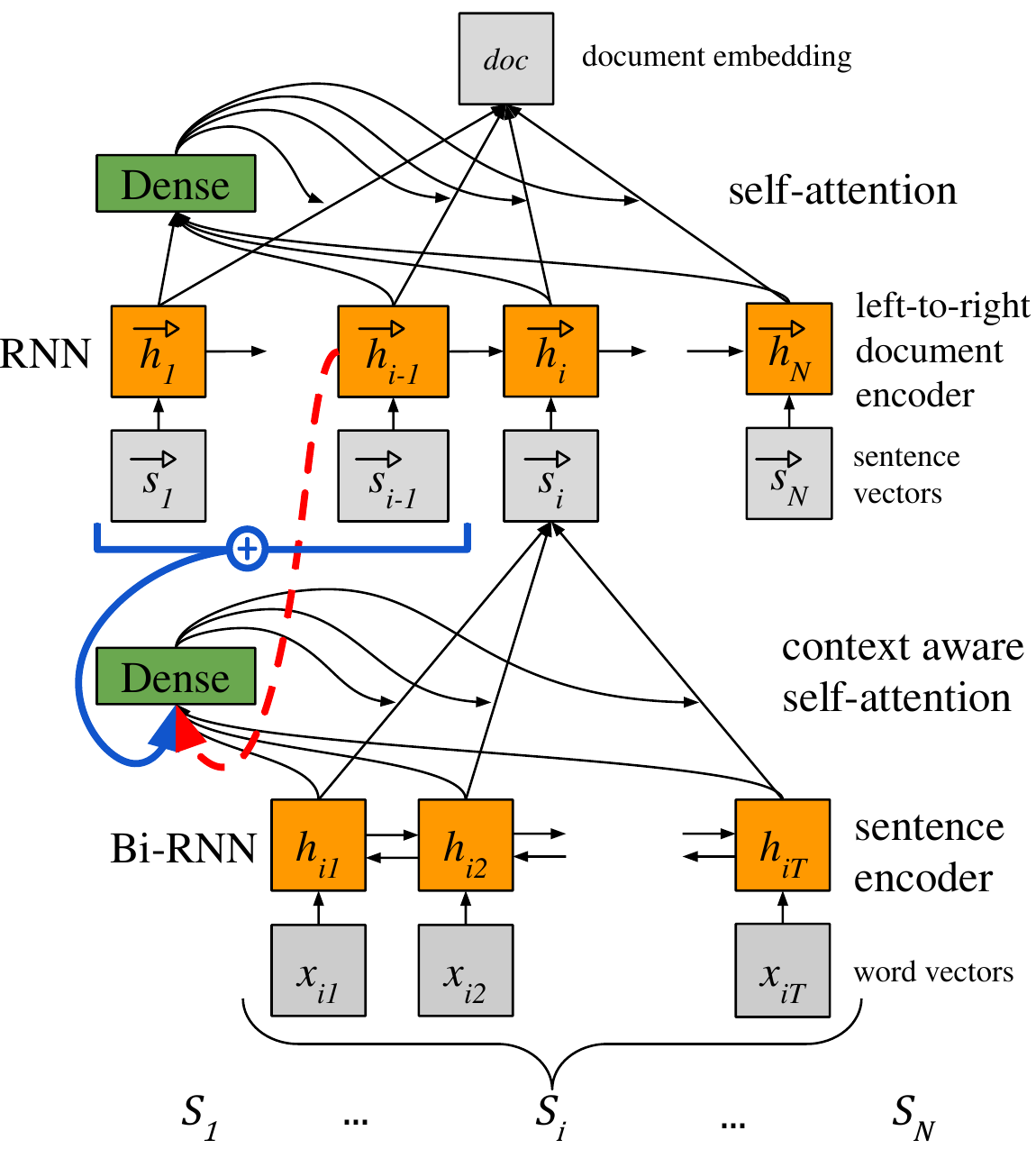}
    \captionsetup{size=small}
    \caption{Our proposed architectural modifications to HAN are shown as the bolded connections.  Blue/continuous: CAHAN-SUM, red/dashed: CAHAN-RNN. Sentence $S_i$ of document $(S_1,\dots,S_N)$ is being encoded. Right-to-left document encoder not shown for clarity. Best seen in color.}\label{fig:CAHAN} 
\end{figure}

\noindent One of the earliest and most influential examples is the Hierarchical Attention Network (HAN) of \citet{yang2016hierarchical} (see Fig. \ref{fig:CAHAN} and section \ref{sec:han}).
It is a two-level architecture, where at level 1, each sentence in the document is separately encoded by the same sentence encoder, resulting in a sequence of sentence vectors. 
That sequence is then processed at level 2 by the document encoder which returns a single vector representing the entire document. 
The sentence and document encoders are both self-attentional bidirectional Recurrent Neural Networks (RNNs), with different parameters.

\paragraph{Observed problem} HAN was highly successful and established new state of the art on six large-scale sentiment and topic classification datasets. 
However, it has a major weakness: at level 1, each sentence is encoded in \textit{isolation}. 
That is, while producing the representation of a given sentence in the document, HAN completely ignores the other sentences. This lack of communication is obviously suboptimal.
For example, in Fig. \ref{fig:example_yelp}, the same highly negative feature (``terrible value'') has been repeated at the beginning of each sentence in the document. 
Because it encodes each sentence independently, HAN has no choice but to spend most of its attentional budget on the most salient feature every time. As a result, HAN neglects the other aspects of the document.
On the other hand, CAHAN is informed about the context, and thus quickly stops spending attention weight on the same highly negative pattern, knowing that is has already been covered.
CAHAN is then able to cover the other topics in the document (``seafood'',``scallops'' and ``mussels''; ``entree'' and ``appetizer''; triple negation in the fourth sentence).

As another example, consider the edge case of a document containing the same sentence repeated several times, as shown in Fig. \ref{fig:example_amazon}.
With HAN, the exact same embedding is produced for each instantiation of the sentence, as a result of the context-blind self-attention mechanism always making the same alignment decisions.
However, the context-aware sentence encoder of CAHAN allows it to extract \textit{complementary}, rather than \textit{redundant} information, from each instantiation of the sentence.
This results in better coverage (``reasonably priced'', ``arrived late''), in a richer document representation, and ultimately in a more accurate prediction (positive instead of very positive).

One may argue that in basic HAN, the document encoder at level 2 already does capture some notion of context, by assigning importance scores to sentences.
However, at level 2, the sentence vectors have already been formed, and it is too late to modify them.
Since the document encoder can only rank the sentence representations, it cannot address issues like high redundancy. In that case, important subtopics or details in the document will not be covered, no matter sentence scores.

\paragraph{Context-aware HAN}

In this work, we propose and evaluate several modifications of the HAN architecture that allow the sentence encoder at level 1 to make its attentional decisions based on contextual information, allowing it to learn richer document representations. Another significant contribution is the introduction of a bidirectional version of the document encoder, where one RNN processes the document forwards, using the preceding sentences as context, and another one processes it backwards, using the following sentences as context.

The remainder of this paper is structured as follows.
We start by formally introducing basic HAN (section \ref{sec:han}), we then explain our contributions (section \ref{sec:cahan}), and detail our experimental setup (section \ref{sec:exps}).
Finally, we interpret our results and list areas of future development (sections \ref{sec:results} and \ref{sec:next}).
Related work is reviewed in section \ref{sec:lit_review}.

\section{HAN}\label{sec:han}
The baseline HAN model as introduced by \citet{yang2016hierarchical} is shown in Fig. \ref{fig:CAHAN} along with our modifications (disregard the bold lines for the baseline).
The sentence and document encoders, used respectively at level 1 and level 2, have different parameters but share the exact same architecture.
Thus, in what follows, we only describe the sentence encoder in detail.

\paragraph{Notation}
Next, we use boldface upper case for tensors, upper case for matrices, boldface lower case for vectors, and lower case for scalars.
We define a document $\mathbf{X} \in \mathbb{R}^{N \times T_i \times d}$ as a sequence of $N$ sentences $(S_1, \dots, S_N)$. Each sentence $S_i$ is a sequence of $T_i$ $d$-dimensional word vectors $(\mathbf{x}_{i1}, \dots, \mathbf{x}_{iT_i}) \in \mathbb{R}^{T_i \times d}$.

\subsection{Sentence encoder}
First, the sentence-level bidirectional RNN $f_s$ processes the input sentence $S_i$ and returns a sequence of $T_i$ $2d_s$-dimensional hidden states $(\mathbf{h}_{i1},\dots, \mathbf{h}_{iT_i}) \in \mathbb{R}^{T_i \times 2d_s}$.
$f_s$ is composed of two non-stacking RNNs $\overrightarrow{f_s}$ and $\overleftarrow{f_s}$ with Gated Recurrent Units \cite{cho2014learning}, respectively parsing $S_i$ from left to right and right to left:

\vspace{-0.5cm}

\begin{align}
\overrightarrow{\mathbf{h}}_{it} &= \overrightarrow{f_s}(\mathbf{x}_{it}, \overrightarrow{\mathbf{h}}_{i,t-1}) \label{eq:han_lr} \\
    \overleftarrow{\mathbf{h}}_{it} &= \overleftarrow{f_s}(\mathbf{x}_{it}, \overleftarrow{\mathbf{h}}_{i,t+1}) \label{eq:han_rl}
\end{align}

\noindent $\overrightarrow{f_s}$ and $\overleftarrow{f_s}$ have the same hidden layer dimensionality $d_s$, but different parameters. At each time step $t$, the word annotations they return are concatenated, producing $2d_s$-dimensional annotations that summarize the immediate context surrounding each word:

\vspace{-0.25cm}

\begin{equation}
\mathbf{h}_{it} =  [\overrightarrow{\mathbf{h}}_{it}; \overleftarrow{\mathbf{h}}_{it}] 
\end{equation}

\noindent Then, a self-attention mechanism computes the representation $\mathbf{s}_i$ of sentence $S_i$ as a weighted sum of its word annotations: 

\vspace{-0.25cm}

\begin{equation}
\mathbf{s}_i = \sum_{t=1}^{T_i} \alpha_{it} \mathbf{h}_{it}  
\end{equation}

\noindent Where the vector of attentional coefficients $\mathbf{\alpha}$ is a softmax-normalized version of the alignment vector $\mathbf{e}$, which itself is obtained by passing the word annotations through a dense layer (parameterized by $W_s \in \mathbb{R}^{2d_s\times 2d_s}$) and comparing the output with a trainable vector $\mathbf{u}_s \in \mathbb{R}^{2d_s}$:

\vspace{-0.2cm}

\begin{equation}\label{eq:self_original}
    e_{it} = \mathbf{u}_s^\top\tanh(W_s \mathbf{h}_{it} + \mathbf{b}_s)
\end{equation}

\noindent $\mathbf{u}_s$ is initialized randomly.
It can be interpreted as a ``super-word'' whose vector contains the ideal combination of latent topics, on average.
The closest the annotation of a word is to this ideal representation, the more attention that word will be given.

The sentence encoder is applied to all sentences in document $\mathbf{X}$, producing a sequence of $N$ sentence vectors $(\mathbf{s_1},\dots,\mathbf{s_N}) \in \mathbb{R}^{N\times 2d_s}$.

\subsection{Document encoder}
The document encoder is a self-attentional bidirectional GRU-RNN, like the sentence encoder, but it has different parameters. The dimensionality of its hidden states is $2d_d$. The document encoder is applied only once, to the sequence of sentence vectors, to produce the sequence of sentence annotations $(\mathbf{h}_{1}, \dots, \mathbf{h}_{N})$. Then, a self-attention layer outputs the final document vector.

\section{Proposed architecture: CAHAN}\label{sec:cahan}
As was previously explained, each sentence is encoded independently by HAN, without considering any kind of contextual information. To solve this issue, we inject a context vector $\mathbf{c_i}$ into the self-attention mechanism, to guide the model during the computation of the word alignment coefficients.
In effect, Eq. \ref{eq:self_original} becomes:

\vspace{-0.5cm}

\begin{equation}\label{eq:self_cahan}
    e_{it} = \mathbf{u}_s^\top\tanh(W_s \mathbf{h}_{it} + W_c\mathbf{c}_i + \mathbf{b}_s)
\end{equation}

\noindent We propose two approaches for computing $\mathbf{c_i}$, namely CAHAN-SUM and CAHAN-RNN, shown as the two bolded connections in Fig. \ref{fig:CAHAN}.

\subsection{Summed context (CAHAN-SUM)}\label{sub:cahan-sum}
We introduce two settings, (1) left-to-right and  bidirectional. Whenever there is no preceding/following sentence, i.e., at the beginning/end of a document, the context vector is initialized with zeroes.

\paragraph{Left-to-right (LR)} In the LR case, the context vector is computed as the sum of the preceding sentence representations:

\vspace{-0.25cm}

\begin{equation}\label{eq:cahan-s-sum}
    \overrightarrow{\mathbf{c}_i} = \sum_{i'=1}^{i-1} \mathbf{s}_{i'}
\end{equation}

\paragraph{Bidirectional (BI)} In the BI case, we compute two context vectors, respectively by summing the representations of the sentences preceding and following the current sentence $S_i$. These two vectors are passed to two identical context-aware self-attention mechanisms (Eq. \ref{eq:self_cahan}) with different parameters.
The resulting forward and backward sentence representations are then processed respectively by the forward and backward RNNs of the document encoder at level 2, and the resulting annotations are concatenated to produce the final sentence annotations.

CAHAN-SUM was inspired by the coverage vectors of seq2seq architectures, which have been shown very effective in addressing under(over)-translation in NMT \cite{tu2016modeling}, and repetition in summarization \cite{see2017get}.
Such coverage vectors are typically computed as the sum, over all previous decoder steps, of the attention distribution over the source tokens. 
However, in our case, we cannot keep track of the attention distribution history, since sentences are unique and cannot be aligned. This is why we work with sentence representations instead.

\paragraph{Centroid version ($\mu$)}
$\overrightarrow{\mathbf{c}_i}$, as defined by Eq. \ref{eq:cahan-s-sum}, grows larger in magnitude as $i$ increases (the sum has more and more terms), which can blur the alignment decisions for the sentences at the end of a document (LR case), or both at the end and beginning of a document, when reading forwards and backwards (BI case). 
Therefore, we also experiment with a centroid, rather than sum, context vector:

\vspace{-0.6cm}

\begin{equation}\label{eq:cahan-s-mean}
    \overrightarrow{\mathbf{c}}_i = \frac{1}{i-1}\sum_{i'=1}^{i-1} \mathbf{s}_{i'}
\end{equation}

\subsection{Recurrent Context (CAHAN-RNN)}\label{sub:cahan-r}
Here, we capitalize on the capability of RNNs, especially when equipped with LSTM or GRU units, to keep track of information over long time periods. 
We simply use as context vector the document encoder annotation at the preceding/following time step. 
That is, we have, in the LR case:

\vspace{-0.5cm}

\begin{equation}
    \overrightarrow{\mathbf{c}_i} = \overrightarrow{\mathbf{h}}_{i-1} 
\end{equation}

\noindent By design,  $\mathbf{h}_{i-1}$ summarizes the entire history $(\mathbf{s_1},\dots,\mathbf{s_{i-1}})$ of sentence vectors, with a preference for the most recent time steps.
If the sequence is very long though, even a GRU-RNN will eventually forget about the first elements.
However, for the relatively short documents we experiment with (see Table \ref{table:datasets}), we can assume the annotations of the document encoder to faithfully represent the entire sequence.

\subsection{Gated context}\label{sub:gated}
In NMT, \citet{tu2017context} introduced a gating mechanism to allow the decoder to balance the contribution of the source and target information in generating the next word.
The same idea can be found in numerous other NMT studies, e.g., \cite{wang2017exploiting,voita2018context,yang2019context}.
Inspired by this line of research, we propose a modification of Eq. \ref{eq:self_cahan} to let our model explicitly decide how much contextual information it should take into account in making its alignment decisions:

\vspace{-0.5cm}

\begin{equation}\label{eq:self_gated}
    e_{it} = \mathbf{u}_s^\top\tanh\big((1-\mathbf{\lambda}) W_s \mathbf{h}_{it} + \mathbf{\lambda} W_c\mathbf{c}_i + \mathbf{b}_s\big)
\end{equation}

\noindent $\mathbf{\lambda}$ is produced by a trainable mechanism taking as input the word annotations and the context vector:

\vspace{-0.25cm}

\begin{equation}\label{eq:lambda}
    \mathbf{\lambda} = \sigma (W_{\lambda_1}\mathbf{h}_{it} + W_{\lambda_2}\mathbf{c}_i + \mathbf{b}_\lambda)
\end{equation}

\noindent The sigmoid activation ensures that $\mathbf{\lambda}$ plays a filtering role, by squashing all its entries to $[0,1]$.

The gate gives more expressiveness to the attention mechanism.
Indeed, contextual information should not always be given the same importance, depending on the situation. E.g., when most of the document has been processed, context is likely to be very important, in order to limit redundancy and increase coverage.
However, at the beginning of a document, or in the case of a very short or focused sentence, context might not be useful as only one single topic might be extractable from the sentence anyways.

From an optimization perspective, $\mathbf{\lambda}$ also has the desirable effect of regulating the magnitude of the context vector, preventing it from pushing the \texttt{tanh} to regions of very small gradient. This is especially useful with CAHAN-SUM, as in that case, $\mathbf{c}_i$ gets large towards the end/beginning of documents (forwards/backwards reading).

\subsection{Complexity and sequentiality}\label{sub:complex}
Assuming that $d \sim 2d_s$ and that $d_s \sim d_d$, which holds in practice under reasonable settings, all matrix multiplications in the network have similar complexity, of order of magnitude $\mathcal{O}(d^2)$.
Moreover, since we use GRU-RNNs, there are 6 matrix multiplication per encoder. This number is doubled, as we use bidirectional RNNs.
Finally, the two self-attention mechanisms, one at each level, add two multiplications\footnote{The dot product comparison operations each have $\mathcal{O}(d)$ complexity, which is negligible.}.
Therefore, in the HAN baseline architecture, there are a total of 26 matrix multiplications (13 at each level).

To that, CAHAN-SUM and CAHAN-RNN simply add one matrix multiplication ($W_c\mathbf{c}_i$ in Eq. \ref{eq:self_cahan}) in the LR case and two in the BI case.
This corresponds to negligible 4\% and 8\% increases in total computational cost.
On top of that, gating adds two multiplications in the LR case ($W_{\lambda_1}\mathbf{h}_{it}$ and $W_{\lambda_2}\mathbf{c}_i$ in Eq. \ref{eq:lambda}) and four in the BI case. All in all, this represents three and six extra multiplications compared to basic HAN, resp. in the LR and BI cases.
Again, this corresponds to small increases in computational cost, of 11.5\% and 23\%, respectively.

However, with CAHAN-SUM, the representations of the preceding/following sentences are now required before computing the current sentence representation. With CAHAN-RNN, one even has to wait until the level 2 RNN has processed the preceding/following sentence vectors before being able to encode the current sentence.
Therefore, the sentence encoding process, which was parallelizable with basic HAN due to independence, has now become a sequential process. This is why in practice, we observe slightly greater runtime increases, in the range 5-22\% (see Table \ref{table:computation_time}).

\section{Experimental setup}\label{sec:exps}

\subsection{Datasets}
We evaluate the quality of the document embeddings learned by the different variants of CAHAN and the HAN baseline on three of the large-scale document classification datasets introduced by
\citet{zhang2015character} and used in the original HAN paper \cite{yang2016hierarchical}. They fall into two categories: 
topic classification (Yahoo) and fine-grained sentiment analysis (Amazon, Yelp).
Dataset statistics are shown in Table \ref{table:datasets}.
Classes are perfectly balanced, for all datasets.

\begin{table*}[ht]
\centering
\scalebox{0.9}{
\begin{tabular}{l|ccccccc}
    \textbf{Dataset} & \textbf{\# Classes} & \textbf{\# Train} & \textbf{\# Test} & \textbf{Avg doc size} & \textbf{Batch size} & \textbf{$d$} & \textbf{Vocab. size} \\
    \hline
    Amazon & 5 & 3,000,000 & 650,000 & 6.2 ($\pm$3) & 128 & 200 & 171,770\\
    Yelp & 5 & 560,000 & 50,000 & 8.8 ($\pm$7.2) & 64 & 100 & 70,240\\
    Yahoo! & 10 & 1,400,000 & 60,000 & 6.7 ($\pm$6.6) & 128 & 200 & 167,933
\end{tabular}
}
\captionsetup{size=footnotesize}
\caption{Statistics and hyperparameters per dataset. Document size is given in number of sentences. Vocabulary size is after preprocessing.}\label{table:datasets}
\end{table*}

\subsection{Model configuration}
This subsection describes the preprocessing and hyperparameter setting we used.

\paragraph{Preprocessing and word embeddings}
For preprocessing (and the HAN baseline), we used the publicly available implementation of \citet{tixier2018notes}\footnote{\scriptsize \url{https://github.com/Tixierae/deep_learning_NLP}}, which closely follows the description and details given in the original HAN paper \cite{yang2016hierarchical}\footnote{The authors did not make their code publicly available and did not reply to private communication.}. More precisely, on each dataset, we randomly split the training set into training (90\%) and validation (10\%). Documents are then tokenized into sentences and sentences are tokenized into tokens. The tokens appearing less than 5 times in the corpus are replaced with a special UNK token. Finally, we pre-train our own word vectors with word2vec \cite{mikolov2013efficient} on the training and validation splits.

\paragraph{Hyperparameters}
We do not tune any hyperparameter except the learning rate (see subsection \ref{sub:training}).
We set the hidden layer dimensionality of the two RNN encoders to $d_s=50$ and $d_d=50$.
Thus, the word annotations, sentence vectors, sentence annotations and document vector all have size 100.

With regularization in mind, we set the dimensionality of the word embeddings to $d=200$ on the very large datasets (Amazon and Yahoo!) and to $d=100$ on Yelp, as shown in Table \ref{table:datasets}. 
We also use a greater batch size of 128 on the large datasets, versus 64 on Yelp.

\subsection{Training details}\label{sub:training}
We zero-pad sentences and documents. Like in \citet{yang2016hierarchical}, to make the most out of each batch, we ensure they are as dense as possible by using a bucketing strategy.
More precisely, we build each batch so that it contains documents of approximately the same size, in number of sentences.
For regularization, we use dropout \cite{srivastava2014dropout} with a rate of 0.5 at each layer.
For classification, the document vectors are passed to a dense layer with softmax activation, whose dimensionality is equal to the number of categories to be predicted.

Initialization has a significant impact on performance.
To make sure the differences we measure are due to differences in the models and not in initial condition, we use the same initialization weights for each model.

\paragraph{SGD with cyclical learning rate}
To minimize the categorical cross-entropy loss, we use the stochastic gradient descent optimizer with a triangular cyclical learning rate schedule and opposite triangular momentum schedule \cite{smith2017cyclical,smith2018disciplined}.
Following the authors' recommendations, we use a fixed $[0.85,0.95]$ momentum range, while for the learning rate, we perform a range test on the validation set, for each model, searching the $[0.001,3]$ range.
With a triangular schedule, the learning rate linearly increases for a certain number of iterations (half-cycle), and then linearly decreases back to its initial value during the second half of the cycle.
Cycles are repeated until training ends.
High learning rate values make training faster, by allowing large updates and the use of greater batch sizes while keeping the amount of regularization constant.
Also, the cyclical schedule injects beneficial stochastic noise to the gradient updates, which improves generalization \cite{jastrzkebski2017three}.

We use cycles of 12 epochs, and an early stopping strategy, monitoring the test loss, with a patience of slightly more than one cycle. We set the maximum number of epochs for all models to 50.

\section{Results}\label{sec:results}

As can be seen in Table \ref{table:res1}, the best version of CAHAN (SUM-BI-$\Sigma$) consistently outperforms the HAN baseline, which shows that taking contextual information into account helps producing better document representations.

\begin{table}[h]
\centering
\scalebox{0.925}{
\begin{tabular}{llccc}
& \textbf{Model} & \textbf{Amazon} & \textbf{Yelp} & \textbf{Yahoo!}\\
\hline
\multirow{ 8}{*}{\rotatebox[origin=c]{90}{{CAHAN-SUM}}} & BI $\Sigma$ & 63.99$^\dagger$ & 66.78$^\dagger$ & \textbf{75.03}$^\dagger$ \\
& \hspace{5pt} + Gate & 63.95$^\dagger$ & \textbf{67.02}$^\dagger$ & 74.98$^\dagger$ \\
 & BI $\mu$ & 63.98$^\dagger$ & 66.90$^\dagger$ & 75.00$^\dagger$ \\
 & \hspace{5pt} + Gate & \textbf{64.10}$^\dagger$ & 66.93$^\dagger$ & 74.98$^\dagger$ \\
 \hdashline
 & LR $\Sigma$ & 63.43 & 66.05 & 74.63\\
 & \hspace{5pt} + Gate & 63.70$^\dagger$ & 66.50 & 74.64\\
 & LR $\mu$ & 63.50 & 66.25 & 74.78\\
 & \hspace{5pt} + Gate & 63.53$^\dagger$ & 66.46 & 74.82\\
\hdashline
& {\small BI-CAHAN-RNN} & 63.17  & 63.35 & 74.46 \\
\hline
& HAN (baseline) & 63.53 & 66.55 & 74.83\\
\end{tabular}
}
\captionsetup{size=footnotesize}
\caption{Classification accuracy. $\mu$ and $\Sigma$ designate the centroid and standard version of CAHAN-SUM (see subsection \ref{sub:cahan-sum}). Best score per column in \textbf{bold}, $^\dagger$better than baseline.}\label{table:res1}
\end{table}

\noindent Also, the two unidirectional variants (LR) slightly underperform the baseline and are clearly inferior to BI, which illustrates the value added by processing the document forwards and backwards, using preceding and following sentences as context.

\paragraph{Summing vs. averaging}
In the unidirectional case, it is surprising to note that CAHAN-SUM-LR-$\mu$ is slightly better than CAHAN-SUM-LR-$\Sigma$, i.e., the centroid-based context vector (Eq. \ref{eq:cahan-s-mean}) is better than the sum-based one (Eq. \ref{eq:cahan-s-sum}).
Indeed, from an information theory standpoint, it should be the opposite, as summing keeps track of all information whereas averaging is lossy.
We hypothesize that towards the end of a document, the sum-based context vector grows large in magnitude, which perturbs the alignment decisions and deteriorates the quality of the sentence vectors.\footnote{This is consistent with the observation that gating (see subsection \ref{sub:gated}) allows CAHAN-SUM-LR-$\Sigma$ to outperform CAHAN-SUM-LR-$\mu$ on 2 datasets out of 3.}
On the other hand, the centroid-based vector, which has constant magnitude, does not suffer from this issue.
We further hypothesize that this issue is attenuated in the bidirectional case (CAHAN-SUM-BI-$\mu$ and CAHAN-SUM-BI-$\Sigma$ are on par) due to a counterbalancing phenomenon.
Indeed, the last sentences processed by the left-to-right encoder are the first ones processed by the right-to-left encoder. 
Therefore, through concatenation, the overall quality of the sentence embeddings stays constant.

\paragraph{Gating} As expected, gating improves performance, especially for the $\Sigma$ variants of CAHAN-SUM (and especially the LR ones). To be noted are significant boosts of 0.45 and 0.24 in accuracy respectively for CAHAN-SUM-LR-$\Sigma$ and CAHAN-SUM-BI-$\Sigma$ on Yelp. On Amazon, gating also offers CAHAN-SUM-LR-$\Sigma$ a nice 0.27 improvement. These positive results give a clue that regulating the magnitude of the context vector $\mathbf{c}_i$ is indeed beneficial.

Nevertheless, gating also improves the performance of the $\mu$ variants of CAHAN, which do not suffer from the context vector magnitude issue. This shows that gating is also helpful via giving more expressiveness to the model. For instance, on Amazon, gating boosts the performance of CAHAN-SUM-BI-$\mu$ by 0.12.

It is interesting to note that overall, gating is mostly effective on Yelp and Amazon. We attribute this to the difference in task. Sentiment analysis may rely more on contextual information than topic classification.

\paragraph{CAHAN-RNN-BI} The consistently bad performance of CAHAN-RNN-BI is to be noted.
This was unexpected, as an equivalent approach was used by \citet{raheja2019dialogue} for dialogue act classification, with significant improvements.
We hypothesize that in our case, CAHAN-RNN-BI is not effective because, unlike utterances in a speech transcription, sentences in a document are not ordered in a temporal fashion.
In other words, sentences far away from the current sentence are not necessarily less relevant than closer sentences.
Thus, considering each sentence equally is better than imposing an implicit time-decay via a RNN.

\paragraph{Runtimes} We compare the average runtime per iteration of some variants of CAHAN to that of basic HAN in Table \ref{table:computation_time}. For CAHAN-SUM-$\Sigma$, we observe that the unidirectional variant (LR) is 5.7\% slower than basic HAN (37 vs. 35ms per iteration), whereas the bidirectional variant (BI) is 23\% slower (43 vs. 35 ms). When gating, these number increase to 14.3\% and 37\% (40 and 48ms vs. 35ms).
These differences are not far from our theoretical expectations (see subsection \ref{sub:complex}), especially for LR. 
Indeed, recall that based on matrix multiplication counts, we had forecasted increases of 4\% and 8\% (11.5\% and 23\% when using gating), respectively for LR and BI.
The gap for BI can be explained by a probable bottleneck in the implementation.

CAHAN-RNN adds the same number of matrix multiplications as CAHAN-SUM, so we should in principle observe the same increases. However, as was explained in subsection \ref{sub:complex}, with CAHAN-RNN we have to wait until the level 2 RNN has processed the preceding or preceding/following sentence vectors (LR or BI case) before being able to encode the current sentence. This explains the extra-time needed (40 vs. 37ms and 49 vs. 43ms).

\begin{table}[h]
\centering
\scalebox{0.85}{
\begin{tabular}{cccc|c}
\multicolumn{2}{c}{\textbf{CAHAN-SUM-$\Sigma$}} & \multicolumn{2}{c}{\textbf{CAHAN-RNN}} & \textbf{HAN} \\
LR & BI & LR & BI & \\
\hline
37 (40) & 43 (48) & 40 & 49 & 35\\
\hline
\end{tabular}
}
\captionsetup{size=footnotesize}
\caption{Average runtime per iteration (in ms), with TensorFlow and a GeForce RTX 2080 Ti GPU, on Amazon. Gating runtimes are within brackets.}\label{table:computation_time}
\end{table}

\section{Related work}\label{sec:lit_review}
In what follows, we provide a review of the relevant literature.
One should note that by context, in this paper, we do not refer to the \textit{intra-sentence} or \textit{internal} context vector of seq2seq encoders \cite{luong2015effective,tu2017context,yang2019context}.
Rather, we refer to the \textit{cross-sentence}, \textit{external}, or \textit{document-level} context.
A few studies only have focused on developing models that take that type of context into account. Most of these studies originate from NMT. We briefly describe them next.

\citet{wang2017exploiting} obtain a global context vector by feeding a fixed number of the previous source sentences to HAN.
They then compare two ways of injecting it into the encoder-decoder model.
First, they propose a \textit{warm-start} approach, in which the encoder and/or decoder hidden states are initialized with the context vector.
Second, they experiment with an \textit{auxiliary} strategy in which the intra-sentence context vector of the encoder is concatenated with the global context vector and passed either (i) directly to the decoder, or (ii) after going through a filtering gate. However, unlike our mechanism and that of \cite{tu2017context,voita2018context,yang2019context}, which all feature two \textit{coupled} gates, the mechanism of \citet{wang2017exploiting} has only one gate.
All strategies proposed by \citet{wang2017exploiting} significantly improve performance, but first place is reached by a combination of the warm-start and gated techniques.

\citet{jean2017does} use an approach similar to the auxiliary approach of \citet{wang2017exploiting}, but they compute the context vector only from the sentence immediately preceding the current source sentence. They then pass it to a dedicated encoder featuring a customized attention mechanism.

\citet{voita2018context} and \citet{zhang2018improving} both extend the Transformer architecture \cite{vaswani2017attention} with a context encoder featuring self-attentional and feed-forward layers. 
Then, \citet{voita2018context} combine the context representation with the source representation produced by the basic Transformer encoder via a gating mechanism.
They do not modify the decoder part of the Transformer.

\citet{zhang2018improving} go one step further by passing the contextual information both to the encoder and the decoder.
In both cases, they add a self-attention mechanism over the context representation.
For the decoder though, they also replace the residual connection after the context self-attention with a gating mechanism, to limit the influence of the context information on the source information.

One piece of work closely related to our study is \citet{celikyilmaz2018deep}. The authors also use a hierarchical attention architecture, where at level 1, each paragraph of a document is encoded by a dedicated encoder.
All encoders share the same stacking bi-RNN architecture.
Moreover, they communicate at each layer to produce context-aware annotations of the words in their paragraphs.
More precisely, at a given layer of the stacking RNN, a given encoder is passed the average of the representations learned by the other encoders at the corresponding layer (like with CAHAN-SUM-$\mu$).
This context vector is then combined with the hidden states and passed as input to the upper RNN layer.
At level 2, the top RNN layer annotations are passed to a word attention mechanism followed by a paragraph attention mechanism.
A major difference with our work is that the authors combine the encoder with a decoder, to perform abstractive summarization of long documents, whereas we only focus on the encoding part.
The word and paragraph attentional decisions at level 2 are thus made by the decoder.
Another significant difference is that the authors use reinforcement learning for training, instead of SGD.

Context-aware models have also been proposed in other NLP domains.
E.g., for spoken language understanding, \citet{shang2019energy} prepend and append the current utterance with two special word vectors respectively summarizing the $C$ preceding and following utterances (respectively), where $C$ is a hyperparameter.
This indirectly initializes the hidden states of the left-to-right and right-to-left components of a bidirectional RNN, like with the warm-start approach of \citet{wang2017exploiting}.
On the other hand, \citet{raheja2019dialogue} rely on a mechanism equivalent to  LR-CAHAN-RNN. They find that it significantly boosts dialogue act classification accuracy. As discussed in section \ref{sec:results}, we hypothesize that CAHAN-RNN is not effective in our application because sentences in a document are not ordered in a temporal manner.

\section{Discussion and next steps}\label{sec:next}
While bidirectional CAHAN-SUM systematically outperforms HAN, margins are modest.
We attribute this to the fact that the datasets used in our experiments contain short documents (see Table \ref{table:datasets}) featuring simple sentences. Thus, the superior expressiveness of CAHAN is not able to show.
To address this issue, we plan in future work to experiment on datasets featuring long documents containing complex sentences.

Moreover, the tasks of sentiment and topic classification do not require a deep understanding of the input documents.
Even if a given document contains some complex sentences with multiple clauses and subtopics, capturing the polarity of only one simple, unambiguous sentence or pattern may be enough to accurately predict the category of the entire document (e.g., ``hated'', ``loved'', ``definitely recommends'', ``will never come back'', etc.).
Thus, we hypothesize that when trained to solve such tasks, CAHAN does not learn to use its context-aware capabilities to the fullest extent.

One solution, and promising area of future work, would consist in explicitly giving CAHAN knowledge about coverage, diversity, and redundancy.
This could be done by modifying the sentence attention mechanism and/or by adding a term to the loss.
Another natural next step is to experiment on tasks requiring a deeper understanding of text, such as end-to-end abstractive summarization.
Some other ideas for improvement include combining CAHAN-SUM with CAHAN-RNN, and/or the mean and centroid vectors; for CAHAN-SUM, obtaining the centroid vector through a trainable mechanism rather than via pooling; and experimenting with a trainable matrix (instead of vector) in the self-attention at both level 1 and level 2, like in \citet{lin2017structured}.
Finally, the context vector could be seen as an external, general summary of the document, and be pre-computed offline by a dedicated encoder.

\section{Conclusion}
In this paper, we proposed several modifications of the HAN architecture that make the sentence encoder context-aware (CAHAN). Results show that taking context into account is beneficial.
Specifically, the bidirectional version of the document encoder, that processes the documents forwards and backwards, using the preceding and following sentences as context, outperforms the HAN baseline on all datasets and is superior to the undirectional variant.
Moreover, the computational overhead is small. Experiments on tasks requiring a deeper understanding of the input documents should better highlight the superiority of CAHAN.

\section{Acknowledgments}
We thank Xiang Zhang for sharing the datasets.
We are grateful to the NVidia corporation for the donation of a GPU as part of their GPU grant program.
This research was supported in part by the open-source project \href{https://linto.ai/}{LinTo}.

\small

\bibliography{mybib}
\bibliographystyle{acl_natbib}

\end{document}